\documentclass[11pt]{article}

\usepackage[preprint]{acl}

\usepackage{times}
\usepackage{latexsym}

\usepackage[T1]{fontenc}
\usepackage[utf8]{inputenc}

\usepackage{microtype}

\usepackage{inconsolata}

\usepackage{graphicx}

\usepackage[table]{xcolor}
\definecolor{oursblue}{RGB}{220,240,255}

\usepackage{amsmath}
\usepackage{amssymb}
\usepackage{booktabs}
\usepackage{multirow}
\usepackage{graphicx}

\usepackage{xcolor}

\usepackage{longtable}
\usepackage{booktabs}
\usepackage{array}
\usepackage{textcomp}

\usepackage{algorithm}
\usepackage{algpseudocode}

\newenvironment{breakablealgorithm}[1]
{%
    \par
    \addvspace{\intextsep}
    \refstepcounter{algorithm}%
    \addcontentsline{loa}{algorithm}
    {\protect\numberline{\thealgorithm}#1}%

    \noindent\rule{\linewidth}{0.8pt}
    \par\nobreak\vspace{2pt}

    \noindent
    \textbf{Algorithm~\thealgorithm: #1}
    \par\nobreak\vspace{2pt}

    \noindent\rule{\linewidth}{0.4pt}
    \par\vspace{3pt}
}
{%
    \par\vspace{2pt}
    \noindent\rule{\linewidth}{0.8pt}
    \par
    \addvspace{\intextsep}
}

\algnewcommand{\LineComment}[1]{%
    \Statex \hspace{\algorithmicindent}
    $\triangleright$ \textit{#1}
}

\usepackage{placeins}

\title{Knowing When to Teach: Learning from Teachable Prefixes \\in Multilingual Reasoning Distillation}
\author{
 \textbf{Jaewon Cho\textsuperscript{1}},
 \textbf{Minseo Yoon\textsuperscript{1}},
 \textbf{Jinyoung Park\textsuperscript{1}},
 \textbf{Jinwoo Seo\textsuperscript{1}},
 \textbf{Jaewon Chu\textsuperscript{2}},
 \\
 \textbf{Alice Oh\textsuperscript{1}$^\dagger$},
 \textbf{Hyunwoo J. Kim \textsuperscript{1}$^\dagger$}
\\
\\
 \textsuperscript{1}KAIST,
 \textsuperscript{2}Korea University
\\
 \small{\texttt{\{cho35750, yms020615, jinyoung.park, sjwoo0612, hyunwoojkim\}@kaist.ac.kr}}\\
 \small{\texttt{alice.oh@kaist.edu}}\\
 \small{\texttt{allonsy07@korea.ac.kr}}
}

\begin{document}
\maketitle
\renewcommand{\thefootnote}{\fnsymbol{footnote}}
\footnotetext[0]{$\dagger$ Corresponding authors.}

\begin{abstract}
On-Policy Self-Distillation (OPSD) improves reasoning by training a student on its own reasoning, using the same model conditioned on reference solutions as a teacher.
Applying OPSD to multilingual reasoning requires the teacher to provide useful feedback across languages, even when the student makes mistakes in intermediate steps.
We observe a phenomenon we call Teacher Drift: as these mistakes accumulate, the teacher may follow the student’s incorrect steps instead of correcting them. Our analysis shows that this shift tends to occur earlier in lower-resource languages.
To account for these differences across languages, we propose Teachable On-Policy Self-Distillation (TOPSD), which determines where and whether teacher supervision should be applied to each reasoning trajectory.
TOPSD uses teacher–student likelihood to estimate where drift begins and select a prefix for supervision. It then verifies whether the teacher can recover the correct answer from that prefix, applying distillation only to prefixes that pass verification. Experiments on multilingual mathematical reasoning across ten languages and three model scales show that TOPSD consistently outperforms conventional OPSD, particularly in low- and medium-resource languages.
\end{abstract}
\section{Introduction}
\begin{figure}[t!]
    \centering
    \includegraphics[width=\columnwidth]{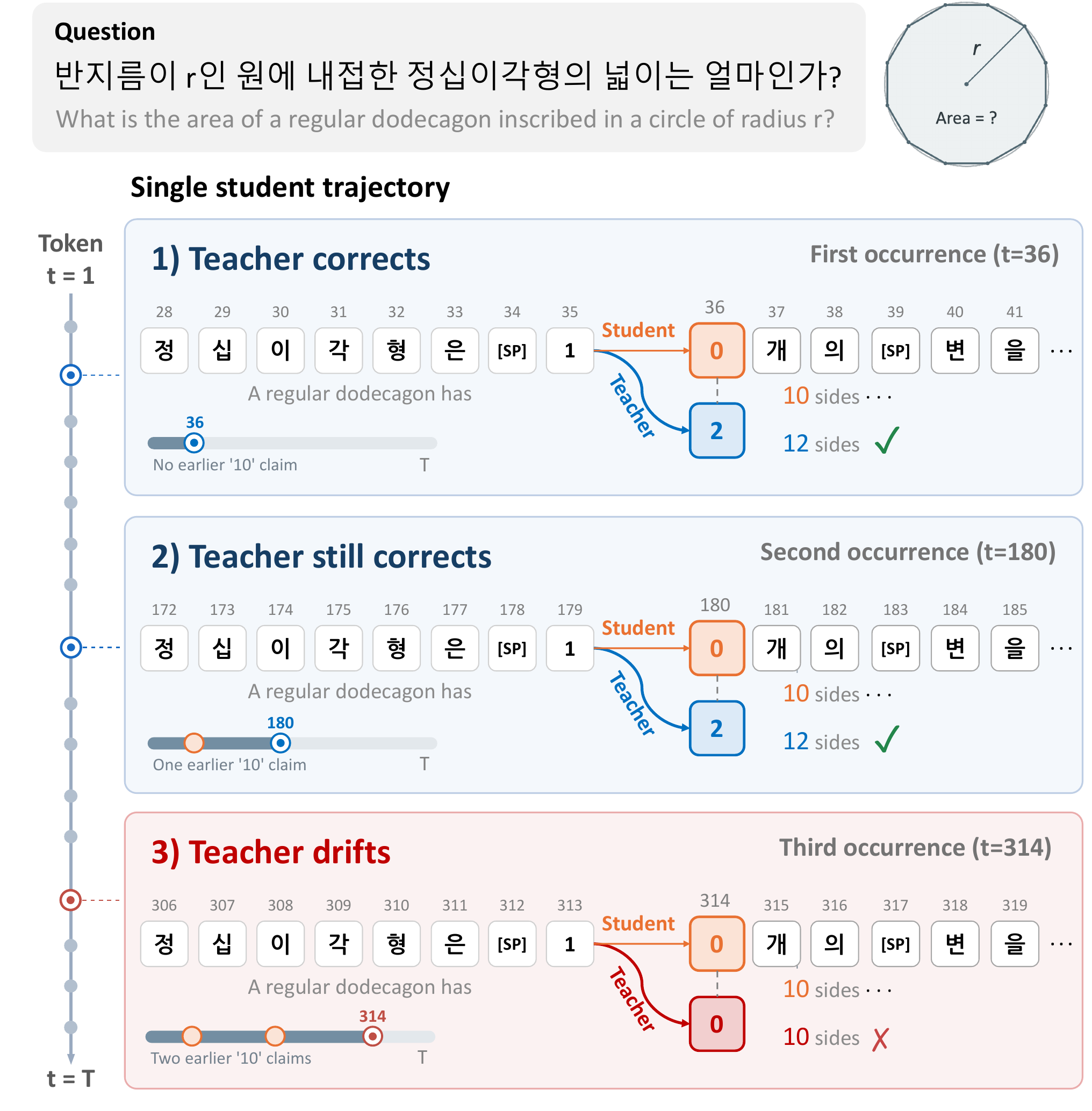}
    \caption{
    \textbf{Teacher Drift in a Korean reasoning example.}
    The teacher's next-token predictions are evaluated on prefixes of one student trajectory.
    At $t=36$ and $180$, it predicts $2$ instead of the student's $0$, correcting a dodecagon's side count from $10$ to $12$; at $t=314$, it predicts $0$, following the repeated error.
    English translations are shown in gray.
    }
    \label{fig:teacher_assimilation}
\end{figure}

Large language models (LLMs)~\citep{brown2020language,achiam2023gpt,team2023gemini,grattafiori2024llama3,yang2024qwen25,guo2025deepseek,yang2025qwen3} have demonstrated strong reasoning capabilities on complex tasks such as mathematical problem solving, logical inference, and complex question answering.
On-Policy Distillation~\citep{ross2011reduction, lin2020autoregressive, agarwal2024policy, zhang2026prefix} further improves these capabilities by training the student with dense token-level supervision on reasoning it generates.
Specifically, the teacher conditions on the student’s prefix, the sequence of tokens generated so far, and uses its next-token distribution to supervise the student.
On-Policy Self-Distillation (OPSD)~\citep{zhao2026self} uses the same model as both student and teacher, with reference solutions provided only to the teacher.
This enables learning from model-generated reasoning with dense teacher feedback, without a separate heavy teacher model~\citep{shenfeld2026self,hubotter2026reinforcement}.

However, extending these benefits to multilingual reasoning is challenging due to the limited reasoning capabilities of the teacher model across languages. 
Since the language model is largely pretrained on high-resource languages such as English, it shows substantially weaker performance in lower-resource languages than in high-resource languages~\citep{zhao2026less}.
Prior work, such as COPSD~\citep{liu2026crosslingual}, addresses it by using English reference solutions to draw on the model’s stronger reasoning ability in English and improve teacher supervision in the target language.

While this effort focuses on strengthening the teacher, it remains underexplored whether the teacher can provide reliable supervision throughout the student’s reasoning.
We therefore investigate how the reliability of teacher supervision changes as the student’s reasoning unfolds across different languages. 
We observe that the teacher may initially correct the student’s errors, but begin to follow the student’s incorrect reasoning as these errors accumulate, even with access to reference solutions. 
We refer to this shift as \textbf{Teacher Drift}.
Figure~\ref{fig:teacher_assimilation} illustrates this behavior in a Korean reasoning example, where the student repeatedly states that a regular dodecagon has ten sides. 
At the first two occurrences, the teacher predicts a token that corrects the number to twelve; at a later occurrence, it predicts the same incorrect token as the student. 
Our multilingual analysis further shows that Teacher Drift tends to emerge earlier in lower-resource languages, suggesting that the teacher’s ability to provide reliable supervision varies across languages.
These observations motivate identifying where the teacher can still provide useful supervision within each reasoning trajectory.

Based on these observations, we propose \textbf{Teachable On-Policy Self-Distillation (TOPSD)}, which determines \emph{where} teacher supervision remains appropriate and \emph{whether} it should be applied.
TOPSD estimates where Teacher Drift begins to select a \emph{teachable prefix}, and then verifies whether the teacher can recover the correct answer from that prefix.
If verification succeeds, distillation is applied to the teachable prefix.

Our contributions are summarized as follows:
\begin{itemize}
    \item We identify \textbf{Teacher Drift}, where the privileged teacher begins to follow the student’s incorrect reasoning as errors accumulate, and characterize its language-dependent patterns through likelihood and recovery analyses.

    \item We propose \textbf{TOPSD}, which combines trajectory-specific prefix selection with teacher recovery verification to determine where and whether to apply supervision.

    \item We demonstrate the effectiveness of TOPSD for multilingual mathematical reasoning across model scales, with improvements particularly in low- and medium-resource languages.
\end{itemize}

\section{Related Work}
\textbf{Multilingual Reasoning.}
Despite advances in LLM reasoning, performance remains uneven across languages, particularly in low-resource settings~\citep{shi2022language,xuan2025mmlu,ghosh2025survey, wang2026polymath}.
Prior work has sought to improve multilingual reasoning through multilingual post-training~\citep{payoungkhamdee2024empirical,chen2024breaking,lai2024mcot,ranaldi2025multilingual,she2024mapo,zhang2026think} and cross-lingual reasoning transfer~\citep{huang2023not,qin2023cross,zhu2024question,tran2025disentangling,barua2026long}.
More recently, COPSD~\citep{liu2026crosslingual} extends on-policy self-distillation to multilingual reasoning by leveraging English problems and reference solutions as privileged teacher information.
Our work complements these efforts by improving the reliability of multilingual on-policy self-distillation, particularly when errors accumulate in student-generated reasoning.

\paragraph{Post-Training for Reasoning.}
Reasoning capabilities are commonly improved through self-training, reinforcement learning, and knowledge distillation.
Self-training bootstraps from model-generated rationales~\citep{zelikman2022star}, while recent post-training methods learn from sampled reasoning trajectories using verifiable outcomes, verifiers, or step-level feedback~\citep{shao2024deepseekmath,trung2024reft,chu2025sft,hosseini2024vstar,wang2024mathshepherd,hwang2024selfexplore,xu2025fullstep}.
Within knowledge distillation~\citep{hinton2015distilling}, GKD and related methods supervise student-generated sequences with teacher predictions~\citep{lin2020autoregressive,gu2024minillm,agarwal2024policy,ko2024distillm,ko2025distillm2}, while OPSD uses the same model as student and privileged teacher under different conditioning contexts~\citep{zhao2026self}.
Recent work further improves the distillation with guided or trust-region rollouts~\citep{li2026rethinking,plyusov2026trust}, selective or adaptive token-level supervision~\citep{wang2026trace,jang2026stable}, and trajectory refinement~\citep{jiang2026trajectory}.
We characterize language-dependent Teacher Drift and verify Teacher Recoverability before distilling selected prefixes of original student trajectories.
\section{Preliminaries}
\label{sec:preliminaries}
\begin{figure*}[t]
    \centering
    \includegraphics[width=0.95\textwidth]{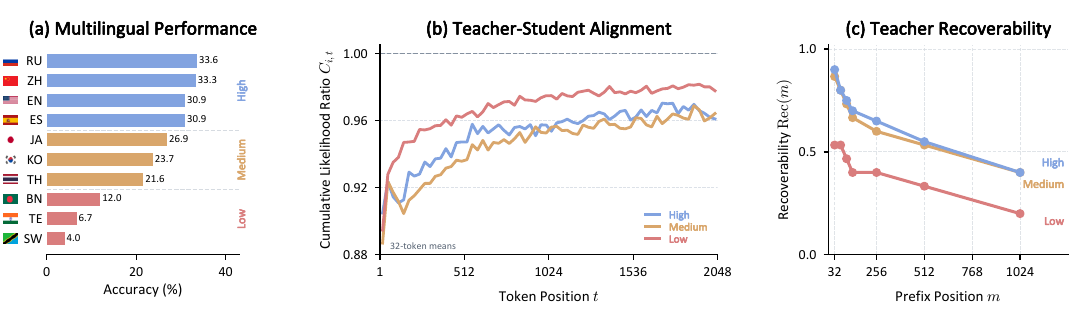}
    \caption{
    \textbf{Teacher Drift in multilingual reasoning.}
    \textbf{(a)} Student accuracy across languages.
    \textbf{(b)} Cumulative teacher-student likelihood ratio $C_{i,t}$ indicates earlier alignment in low-resource languages.
    \textbf{(c)} Teacher recoverability $\mathrm{Rec}(m)$ decreases with longer student prefixes.
    Colors indicate language-resource groups.
    }
    \label{fig:figure2}
\end{figure*}
\begin{figure*}[t]
    \centering
    \includegraphics[width=0.95\textwidth]{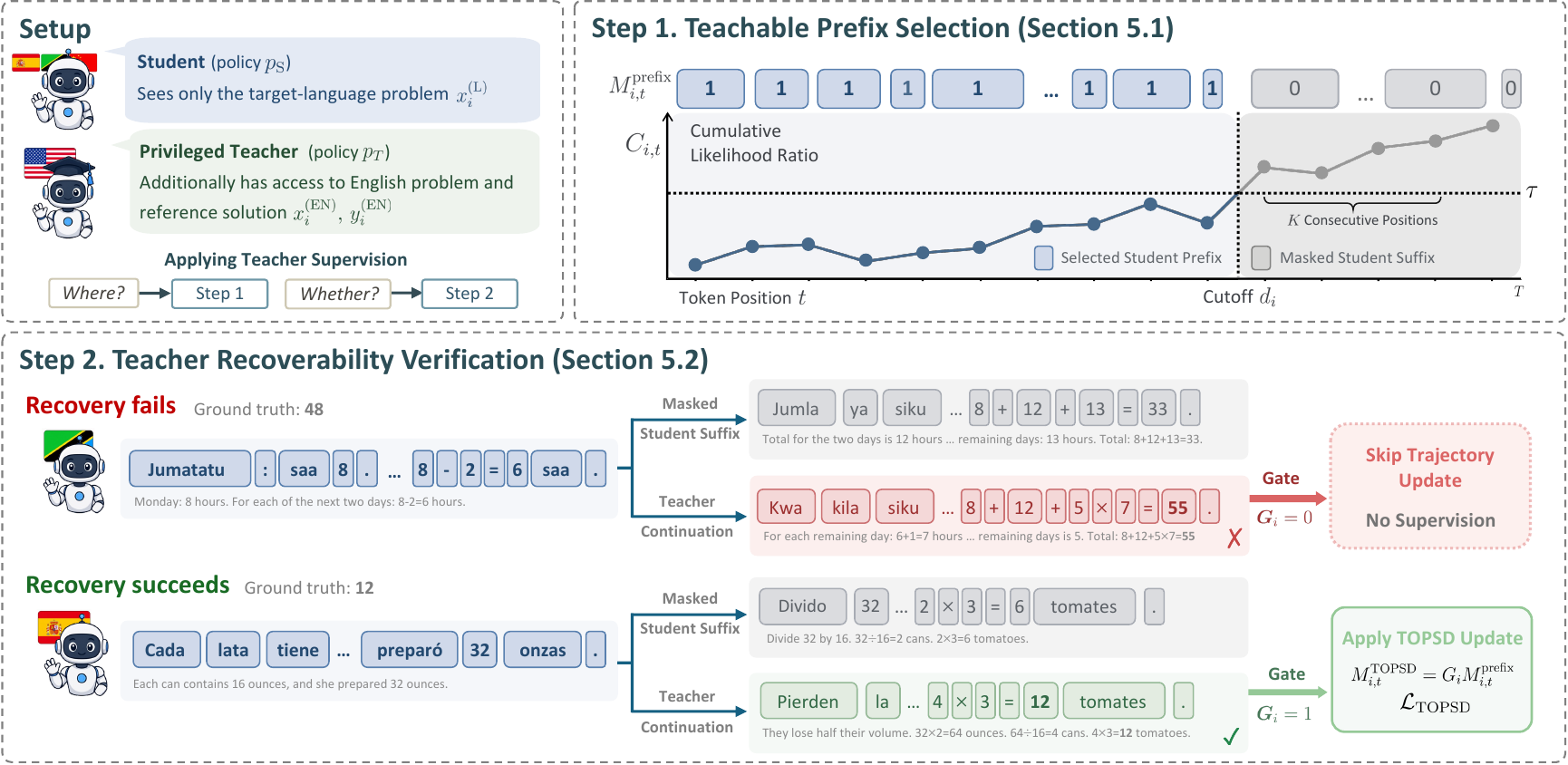}
    \caption{
    \textbf{Overview of TOPSD.}
    The student generates a reasoning trajectory in the target language, while the privileged teacher additionally observes the English problem and reference solution.
    \textbf{Step 1 (Teachable Prefix Selection)}
    uses the cumulative teacher-student likelihood ratio to select a candidate teachable prefix, determining \emph{where} teacher supervision may be applied.
    \textbf{Step 2 (Teacher Recoverability Verification)}
    checks whether the teacher can recover the correct answer from the selected prefix, determining \emph{whether} supervision should be applied.
    Only prefixes that pass verification receive distillation on the corresponding states of the original student trajectory.
    English translations are shown in gray.
    }
    \label{fig:figure3}
\end{figure*}
On-Policy Self-Distillation (OPSD)~\citep{zhao2026self} improves reasoning by training a student on its own generated trajectories with dense supervision from the same model conditioned on reference solutions, without requiring a separate teacher model.
Let $\mathcal{D}$ denote a reasoning dataset consisting of a set of input problems $x$ and their reference solutions $y^\star$.
Rather than relying on a separate teacher model, OPSD instantiates both the student and the privileged teacher from the same underlying language model $\pi_\theta$, while conditioning them on different information.
The student $p_{\mathrm{S}}$ observes only the problem $x$, whereas the privileged teacher $p_{\mathrm{T}}$ additionally observes the reference solution $y^\star$:
\begin{equation}
\begin{aligned}
p_{\mathrm{S}}(\cdot\mid x)
&\triangleq \pi_{\theta}(\cdot\mid x),\\
p_{\mathrm{T}}(\cdot\mid x,y^\star)
&\triangleq \pi_{\theta}(\cdot\mid x,y^\star).
\end{aligned}
\label{eq:opsd_teacher_dist}
\end{equation}
For each problem $x$, the student first samples a trajectory
\begin{equation}
\hat{y} = \left(\hat{y}_1, \dots, \hat{y}_{\left\lvert \hat{y} \right\rvert}  \right)
\sim
p_{\mathrm{S}}(\cdot\mid x).
\label{eq:opsd_rollout}
\end{equation}
Then, OPSD trains the student by minimizing the token-level distributional discrepancy between the two policies along the student-generated trajectory:

{
\small
\begin{equation}
\mathcal{L}_{\mathrm{OPSD}}
=
\mathbb{E}_{(x,y^\star)\sim\mathcal{D}}
\mathbb{E}_{\hat{y}\sim p_{\mathrm{S}}(\cdot\mid x)}
\left[
\frac{1}{|\hat{y}|}
\sum_{t=1}^{|\hat{y}|}
D\!\left(p_{\mathrm{S},t},p_{\mathrm{T},t}\right)
\right],
\label{eq:opsd}
\end{equation}
}
where
$p_{\mathrm{S},t}
:= p_{\mathrm{S}}(\cdot\mid x,\hat{y}_{<t})$
and
$p_{\mathrm{T},t}
:= p_{\mathrm{T}}(\cdot\mid x,y^\star,\hat{y}_{<t})$
are the student and teacher next-token distributions conditioned on the same student-generated prefix, and $D(\cdot,\cdot)$ denotes a distributional divergence.

The simple way to design a teacher prompt in multilingual settings is to use a reference solution $y^{(L)}$ for each target language $L$~(\textit{i.e., }$y^\star \triangleq y^{(L)}$).
However, the model may find it harder to make use of a reasoning trace written in a non-English language, since most of its reasoning ability is learned from English data.
So, COPSD~\citep{liu2026crosslingual} provides the English question $x^{(\mathrm{EN})}$ and its reference solution $y^{(\mathrm{EN})}$ as the privileged knowledge $y^\star$ to leverage the reasoning capabilities of LLMs.
The teacher prompt for target language $L$ can be written as:
$y^\star
\triangleq
\left(x^{(\mathrm{EN})},
y^{(\mathrm{EN})}\right)$.

Even though OPSD variants have shown their success in multiple reasoning tasks, it is still under-explored in multilingual settings.
This is particularly challenging because OPSD relies on the teacher to provide reliable supervision over student-generated trajectories.
In multilingual settings, this supervision faces two potential challenges.
First, errors in student-generated prefixes may make it difficult for the teacher to provide corrective guidance.
Second, the teacher itself may have limited reasoning capabilities in non-English languages, where training data is often less abundant.
These challenges raise concerns about whether the teacher can consistently provide reliable supervision throughout student rollouts.
\section{Observation}
\label{sec:crpd}

\subsection{Observing Teacher Drift}
\label{sec:teacher-drift}
We investigate how the privileged teacher's behavior changes as the student-generated prefix grows, and whether these changes differ across languages.
In particular, we examine whether the teacher
increasingly follows the student-generated trajectory and whether this alignment is accompanied by a reduced ability to correct erroneous reasoning.

\paragraph{Cumulative Teacher-Student Likelihood Ratio.}
We first examine how the teacher's likelihood assignments to student-generated tokens change as the student prefix grows.
For each language $L$, let $i$ index problem instances and $\hat{y}_i^{(L)}$ denote the corresponding student-generated trajectory.
We define a cumulative teacher-student likelihood ratio $C_{i,t}$ to measure how producible the student-generated tokens are from the teacher.

For each prefix position $t>1$, we define the cumulative teacher-student likelihood ratio as 
\begin{equation}
C_{i,t}
=
\exp\!\left(
\frac{1}{t-1}
\sum_{s=1}^{t-1} N_{i,s}
\right),
\end{equation}
where
$N_{i,t}
=
\log\!\bigl[
p_{\mathrm{T},i,t}(\hat{y}_{i,t}^{(L)})
/
p_{\mathrm{S},i,t}(\hat{y}_{i,t}^{(L)})
\bigr]$
denotes the teacher-student log-likelihood gap at token $t$.
Here $p_{\mathrm{S},i,t}$ and $p_{\mathrm{T},i,t}$ denote the student and privileged teacher next-token distributions under their respective prompts and the same student prefix $\hat{y}_{i,<t}^{(L)}$.
It measures the geometric mean of the teacher-to-student likelihood ratios on the observed student prefix.
As $C_{i,t}$ approaches $1$, the average teacher-student log-likelihood gap on the observed student prefix approaches zero.

\paragraph{Language-Dependent Teacher Drift.}
To examine how the teacher's behavior changes over student-generated trajectories, we analyze the cumulative teacher-student likelihood ratio across prefix lengths and language-resource groups.
Figure~\ref{fig:figure2}(b) shows that $C_t$ generally increases toward $1$ as the student prefix grows, indicating that the teacher and student assign increasingly similar relative likelihoods to the observed student-generated tokens.
Notably, this trend emerges earlier and is more pronounced in low-resource languages, suggesting that teacher is influenced by the student at different rates across languages.

\paragraph{Degradation of Teacher Recoverability.}
To assess the teacher's ability to correct erroneous student reasoning, we evaluate whether it can recover the correct answer from student-generated prefixes.
We consider a set of examples $\mathcal{I}$ where the student answers incorrectly but the privileged teacher solves the problem correctly without a student-generated prefix.

For each prefix position $m$, we condition the
privileged teacher on the first $m-1$
student-generated tokens and generate a continuation:
\begin{equation}
\begin{aligned}
z_{i,\geq m}^{(L)}
=
\operatorname{GreedyDecode}
\!\left(
p_{\mathrm{T}},
x_i^{(L)}, y_i^\star,\hat{y}^{(L)}_{i,<m}
\right)\!.
\end{aligned}
\label{eq:teacher_continuation}
\end{equation}
We measure \emph{teacher recoverability} as
the fraction of examples for which the teacher
reaches the correct answer from the given prefix:
\begin{equation}
\operatorname{Rec}(m)
=
\frac{1}{|\mathcal{I}|}
\sum_{i\in\mathcal{I}}
\mathcal{V}
\left(
\hat{y}^{(L)}_{i,<m}
\oplus z_{i,\geq m}^{(L)},
a_i
\right),
\end{equation}
where $a_i$ is the ground-truth answer to the $i$-th problem, $\mathcal{V}$ evaluates answer correctness, and $\oplus$ denotes sequence concatenation.

As shown in Figure~\ref{fig:figure2}(c),
teacher recoverability decreases as the student
prefix grows.
The teacher increasingly fails to reach the correct answer when conditioned on longer student-generated prefixes.
This means that the teacher may lose its correct reasoning capability as the length of student-generated prefix is longer. 

Together, these findings show that increasing
teacher-student likelihood alignment is accompanied by declining teacher recoverability, while the alignment patterns differ across languages.
These observations show that the reliability of teacher supervision may depend on the student-generated prefix, motivating the identification of \emph{teachable prefixes} rather than applying supervision uniformly throughout the student trajectory.

\section{Teachable On-Policy Self-Distillation}
To address limitations of OPSD in multilinguality, such as Teacher Drift and degradation
of teacher recoverability, we propose \emph{Teachable On-Policy Self-Distillation} (TOPSD), which determines \emph{where} and \emph{whether} to apply teacher supervision along student-generated trajectories.
As illustrated in Figure~\ref{fig:figure3}, TOPSD consists of two stages: Teachable prefix selection and Teacher Recoverability Verification.
\textbf{Teachable prefix selection} uses the cumulative teacher-student likelihood ratio to identify a teachable prefix before sustained teacher drift.
\textbf{Teacher recoverability verification} then evaluates whether the teacher can recover the correct answer from the selected prefix.
Only when verification succeeds does TOPSD apply distillation to the selected states of the original student trajectory to provide the meaningful guidance.

\subsection{Teachable Prefix Selection}
\label{sec:teachable-prefix}
As observed in Section~\ref{sec:teacher-drift},
Teacher Drift manifests as increasing likelihood alignment between the teacher and student along student-generated trajectories.
To avoid applying supervision to later states where the teacher may increasingly follow the student's reasoning, we select a candidate teachable prefix based on the cumulative teacher-student likelihood ratio $C_{i,t}$.

Specifically, we identify the earliest position $d_i$
at which $C_{i,t}$ exceeds a threshold $\tau$
for $K$ consecutive positions, which determines the cutoff for prefix selection:
\begin{equation}
d_i
=
\min\!\left\{
j \;\middle|\;
C_{i,j+k}>\tau,\;
\forall k\in\{0,\ldots,K-1\}
\right\}.
\end{equation}
We use consecutive threshold crossings to identify sustained teacher-student likelihood alignment rather than isolated fluctuations.

The corresponding prefix selection mask is
$M_{i,t}^{\mathrm{prefix}}=\mathbb{I}[t<d_i]$,
such that only states within the selected teachable prefix
are eligible for teacher supervision.
If no valid position is detected, we set $d_i=T_i+1$, where $T_i$ is the trajectory length, retaining the entire trajectory as a candidate prefix.
The selected prefix is then evaluated through Teacher Recoverability Verification.

\subsection{Teacher Recoverability Verification}
\label{sec:teacher-recoverability-verification}
To ensure that the selected prefixes provide reliable teacher supervision, we further assess whether the privileged teacher can successfully recover the correct answer from each prefix.
Specifically, we introduce \emph{Teacher Recoverability Verification}, which verifies the teacher's ability to reach the correct solution before using its predictions for distillation. 

Given the selected prefix $\hat{y}_{i,<d_i}^{(L)}$,
we generate a teacher continuation using greedy decoding
as in Eq.~\eqref{eq:teacher_continuation}, with $m=d_i$.
We then concatenate the teacher continuation with the selected
student prefix and evaluate whether the resulting trajectory
reaches the correct answer:
\begin{equation}
G_i
=
\mathcal{V}\!\left(
\hat{y}_{i,<d_i}^{(L)}
\oplus z_{i,\geq d_i}^{(L)},
a_i
\right)
\in\{0,1\},
\end{equation}
where $\mathcal{V}$ denotes the answer verifier
and $a_i$ is the ground-truth answer.

We apply teacher supervision only when the teacher successfully recovers the correct answer ($G_i=1$).
Combining this verification result with the prefix selection mask, we define the final supervision mask as
\begin{equation}
M_{i,t}^{\mathrm{TOPSD}}
=
G_i M_{i,t}^{\mathrm{prefix}}.
\end{equation}
Importantly, the teacher continuation is used only for verification, not as a training target.
Distillation is performed on the selected states of the original student-generated trajectory, preserving the on-policy nature of TOPSD.

\subsection{TOPSD Objective}
\label{sec:topsd_objective}
TOPSD follows the token-level distillation framework of OPSD while restricting supervision using the final mask $M_{i,t}^{\mathrm{TOPSD}}$.
To prevent longer selected prefixes from dominating the training objective and to balance supervision across languages, we normalize the loss at the trajectory and language levels.

Let $n_i = \sum_{t=1}^{T_i} M_{i,t}^{\mathrm{TOPSD}}$ denote the number of supervised tokens in trajectory $i$.
We define the set of eligible trajectories in language $\ell$ as $\mathcal{B}_{\ell}^{+} = \{i \in \mathcal{B} \mid \ell_i = \ell,\; n_i > 0\}$, and the set of languages with eligible trajectories as $\mathcal{A} = \{\ell \mid |\mathcal{B}_{\ell}^{+}| > 0\}$.
The resulting TOPSD objective is
{\small
\begin{equation}
\mathcal{L}_{\mathrm{TOPSD}}
=
\mathbb{E}_{\ell}
\left[
\mathbb{E}_{i\mid\ell}
\left[
\frac{1}{n_i}
\sum_{t=1}^{T_i}
M_{i,t}^{\mathrm{TOPSD}}
D\!\left(p_{\mathrm{S},i,t},p_{\mathrm{T},i,t}\right)
\right]
\right],
\label{eq:topsd_objective}
\end{equation}
}
where
$\mathbb{E}_{\ell}[\cdot]
:= \frac{1}{|\mathcal{A}|}
\sum_{\ell\in\mathcal{A}}[\cdot]$
and
$\mathbb{E}_{i\mid\ell}[\cdot]
:= \frac{1}{|\mathcal{B}_{\ell}^{+}|}
\sum_{i\in\mathcal{B}_{\ell}^{+}}[\cdot]$
denote uniform averaging over eligible languages and eligible trajectories within each language, respectively.
\section{Experiments}

\subsection{Experimental Settings}
\begin{table*}[t]
\centering
\footnotesize
\setlength{\tabcolsep}{2.0pt}
\renewcommand{\arraystretch}{1.03}
\setlength{\extrarowheight}{0.6pt}
\resizebox{0.95\textwidth}{!}{%
\begin{tabular}{@{}cll|ccc|c|ccc|c|cccc|c|c@{}}
\toprule
Model & Metric & Method
& \multicolumn{4}{c|}{Low}
& \multicolumn{4}{c|}{Medium}
& \multicolumn{5}{c|}{High}
& Overall \\
\cmidrule(lr){4-7}
\cmidrule(lr){8-11}
\cmidrule(lr){12-16}
\cmidrule(lr){17-17}
& &
& BN & TE & SW & Avg
& JA & KO & TH & Avg
& EN & ES & RU & ZH & Avg
& Avg \\
\midrule

% =========================================================
% Qwen3-1.7B
% =========================================================

\multirow{8}{*}{\textit{\textbf{Qwen3-1.7B}}}
& \multirow{4}{*}{Avg@3}
& Vanilla
& 2.13 & 0.00 & 0.27 & 0.80
& \underline{11.47} & 6.40 & 6.13 & 8.00
& 17.87 & \underline{15.73} & \underline{14.93} & 17.33 & 16.47
& 9.23 \\

&
& OPSD
& 2.93 & \underline{1.07} & \textbf{1.60} & \underline{1.87}
& 10.13 & 6.40 & 6.93 & 7.82
& \textbf{21.33} & 14.67 & 12.53 & 18.40 & 16.73
& 9.60 \\

&
& COPSD
& \underline{3.47} & \underline{1.07} & \underline{1.07} & \underline{1.87}
& 9.87 & \underline{8.53} & \underline{8.27} & \underline{8.89}
& \underline{20.27} & 13.87 & 13.07 & \underline{20.00} & \underline{16.80}
& \underline{9.95} \\

&
& \cellcolor{oursblue}\textbf{TOPSD}
& \cellcolor{oursblue}\textbf{4.53}
& \cellcolor{oursblue}\textbf{2.13}
& \cellcolor{oursblue}\textbf{1.60}
& \cellcolor{oursblue}\textbf{2.76}
& \cellcolor{oursblue}\textbf{13.07}
& \cellcolor{oursblue}\textbf{10.13}
& \cellcolor{oursblue}\textbf{9.87}
& \cellcolor{oursblue}\textbf{11.02}
& \cellcolor{oursblue}\underline{20.27}
& \cellcolor{oursblue}\textbf{17.07}
& \cellcolor{oursblue}\textbf{17.07}
& \cellcolor{oursblue}\textbf{20.27}
& \cellcolor{oursblue}\textbf{18.67}
& \cellcolor{oursblue}\textbf{11.60} \\

\cmidrule(lr){2-17}

&
\multirow{4}{*}{Pass@3}
& Vanilla
& 4.80 & 0.00 & 0.80 & 1.87
& 18.40 & 11.20 & 10.40 & 13.33
& 26.40 & \textbf{28.00} & \textbf{25.60} & 28.80 & \underline{27.20}
& 15.44 \\

&
& OPSD
& 4.80 & \underline{3.20} & \textbf{4.00} & \underline{4.00}
& \underline{21.60} & 12.00 & \underline{12.80} & \underline{15.47}
& \underline{30.40} & \textbf{28.00} & 21.60 & 28.00 & 27.00
& \underline{16.64} \\

&
& COPSD
& \underline{6.40} & 2.40 & \underline{3.20} & \underline{4.00}
& 18.40 & \underline{15.20} & \underline{12.80} & \underline{15.47}
& \textbf{32.00} & 23.20 & 21.60 & \underline{30.40} & 26.80
& 16.56 \\

&
& \cellcolor{oursblue}\textbf{TOPSD}
& \cellcolor{oursblue}\textbf{8.00}
& \cellcolor{oursblue}\textbf{4.80}
& \cellcolor{oursblue}\underline{3.20}
& \cellcolor{oursblue}\textbf{5.33}
& \cellcolor{oursblue}\textbf{23.20}
& \cellcolor{oursblue}\textbf{17.60}
& \cellcolor{oursblue}\textbf{15.20}
& \cellcolor{oursblue}\textbf{18.67}
& \cellcolor{oursblue}\textbf{32.00}
& \cellcolor{oursblue}\underline{27.20}
& \cellcolor{oursblue}\underline{24.80}
& \cellcolor{oursblue}\textbf{31.20}
& \cellcolor{oursblue}\textbf{28.80}
& \cellcolor{oursblue}\textbf{18.72} \\

\addlinespace[2pt]
\midrule
\addlinespace[1pt]

% =========================================================
% Qwen3-4B
% =========================================================

\multirow{8}{*}{\textit{\textbf{Qwen3-4B}}}
& \multirow{4}{*}{Avg@3}
& Vanilla
& 5.60 & 2.67 & 1.87 & 3.38
& \underline{17.87} & \underline{18.40} & \underline{15.47} & \underline{17.24}
& 21.87 & 23.20 & 26.13 & 21.87 & 23.27
& 15.49 \\

&
& OPSD
& \underline{6.93} & \underline{3.20} & \underline{2.13} & \underline{4.09}
& 17.60 & 16.53 & 13.07 & 15.73
& \textbf{27.47} & 25.33 & \underline{26.93} & \textbf{28.00} & \textbf{26.93}
& \underline{16.72} \\

&
& COPSD
& 5.33 & 2.93 & 1.60 & 3.29
& 17.07 & 15.73 & 12.00 & 14.93
& \underline{27.20} & \textbf{26.13} & 26.67 & \underline{27.20} & \underline{26.80}
& 16.19 \\

&
& \cellcolor{oursblue}\textbf{TOPSD}
& \cellcolor{oursblue}\textbf{8.00}
& \cellcolor{oursblue}\textbf{3.73}
& \cellcolor{oursblue}\textbf{3.20}
& \cellcolor{oursblue}\textbf{4.98}
& \cellcolor{oursblue}\textbf{19.73}
& \cellcolor{oursblue}\textbf{21.07}
& \cellcolor{oursblue}\textbf{17.33}
& \cellcolor{oursblue}\textbf{19.38}
& \cellcolor{oursblue}25.07
& \cellcolor{oursblue}\underline{25.87}
& \cellcolor{oursblue}\textbf{27.47}
& \cellcolor{oursblue}26.40
& \cellcolor{oursblue}26.20
& \cellcolor{oursblue}\textbf{17.79} \\

\cmidrule(lr){2-17}

&
\multirow{4}{*}{Pass@3}
& Vanilla
& 11.20 & 4.00 & 4.80 & 6.67
& \underline{23.20} & \underline{25.60} & \underline{23.20} & 24.00
& 30.40 & 32.00 & 35.20 & 29.60 & 31.80
& 21.92 \\

&
& OPSD
& \underline{12.00} & \textbf{8.00} & \underline{5.60} & \underline{8.53}
& \textbf{27.20} & 24.80 & 20.00 & 24.00
& \textbf{38.40} & \underline{38.40} & 38.40 & \textbf{38.40} & \textbf{38.40}
& \underline{25.12} \\

&
& COPSD
& 10.40 & 6.40 & 4.80 & 7.20
& \textbf{27.20} & 24.80 & 21.60 & \underline{24.53}
& \textbf{38.40} & \textbf{39.20} & \textbf{40.00} & 36.00 & \textbf{38.40}
& 24.88 \\

&
& \cellcolor{oursblue}\textbf{TOPSD}
& \cellcolor{oursblue}\textbf{13.60}
& \cellcolor{oursblue}\underline{7.20}
& \cellcolor{oursblue}\textbf{7.20}
& \cellcolor{oursblue}\textbf{9.33}
& \cellcolor{oursblue}\textbf{27.20}
& \cellcolor{oursblue}\textbf{30.40}
& \cellcolor{oursblue}\textbf{27.20}
& \cellcolor{oursblue}\textbf{28.27}
& \cellcolor{oursblue}\underline{34.40}
& \cellcolor{oursblue}36.80
& \cellcolor{oursblue}\underline{39.20}
& \cellcolor{oursblue}\underline{36.80}
& \cellcolor{oursblue}\underline{36.80}
& \cellcolor{oursblue}\textbf{26.00} \\

\addlinespace[2pt]
\midrule
\addlinespace[1pt]

% =========================================================
% Qwen3-8B
% =========================================================

\multirow{8}{*}{\textit{\textbf{Qwen3-8B}}}
& \multirow{4}{*}{Avg@3}
& Vanilla
& \underline{12.00} & \underline{6.67} & \underline{4.00} & \underline{7.56}
& \underline{26.93} & \underline{23.73} & \underline{21.60} & \underline{24.09}
& 30.93 & \underline{30.93} & 33.60 & 33.33 & 32.20
& 22.37 \\

&
& OPSD
& 11.20 & 5.07 & 3.20 & 6.49
& 25.87 & 20.53 & 18.40 & 21.60
& \textbf{37.87} & 28.53 & \underline{34.67} & \underline{38.40} & 34.87
& 22.37 \\

&
& COPSD
& 9.60 & 5.07 & 1.87 & 5.51
& 26.13 & 21.87 & 18.93 & 22.31
& \textbf{37.87} & 29.87 & 34.40 & \textbf{39.20} & \underline{35.33}
& \underline{22.48} \\

&
& \cellcolor{oursblue}\textbf{TOPSD}
& \cellcolor{oursblue}\textbf{13.60}
& \cellcolor{oursblue}\textbf{7.73}
& \cellcolor{oursblue}\textbf{5.60}
& \cellcolor{oursblue}\textbf{8.98}
& \cellcolor{oursblue}\textbf{30.13}
& \cellcolor{oursblue}\textbf{27.47}
& \cellcolor{oursblue}\textbf{25.07}
& \cellcolor{oursblue}\textbf{27.56}
& \cellcolor{oursblue}\underline{36.00}
& \cellcolor{oursblue}\textbf{31.73}
& \cellcolor{oursblue}\textbf{36.27}
& \cellcolor{oursblue}37.60
& \cellcolor{oursblue}\textbf{35.40}
& \cellcolor{oursblue}\textbf{25.12} \\

\cmidrule(lr){2-17}

&
\multirow{4}{*}{Pass@3}
& Vanilla
& \underline{20.00} & \underline{11.20} & \underline{8.00} & \underline{13.07}
& \underline{39.20} & \underline{36.00} & \underline{32.00} & \underline{35.73}
& 39.20 & 40.80 & 42.40 & 42.40 & 41.20
& 31.12 \\

&
& OPSD
& 16.80 & 8.00 & 6.40 & 10.40
& \underline{39.20} & 31.20 & 29.60 & 33.33
& \underline{48.00} & \underline{42.40} & \textbf{47.20} & \textbf{50.40} & \textbf{47.00}
& \underline{31.92} \\

&
& COPSD
& 16.80 & 8.80 & 4.80 & 10.13
& \underline{39.20} & 33.60 & 27.20 & 33.33
& \textbf{48.80} & 41.60 & \underline{46.40} & \textbf{50.40} & \underline{46.80}
& 31.76 \\

&
& \cellcolor{oursblue}\textbf{TOPSD}
& \cellcolor{oursblue}\textbf{22.40}
& \cellcolor{oursblue}\textbf{12.00}
& \cellcolor{oursblue}\textbf{12.80}
& \cellcolor{oursblue}\textbf{15.73}
& \cellcolor{oursblue}\textbf{43.20}
& \cellcolor{oursblue}\textbf{40.00}
& \cellcolor{oursblue}\textbf{36.00}
& \cellcolor{oursblue}\textbf{39.73}
& \cellcolor{oursblue}46.40
& \cellcolor{oursblue}\textbf{44.00}
& \cellcolor{oursblue}\underline{46.40}
& \cellcolor{oursblue}\underline{49.60}
& \cellcolor{oursblue}46.60
& \cellcolor{oursblue}\textbf{35.28} \\

\bottomrule
\end{tabular}%
}

\caption{
Avg@3 and Pass@3 (\%) on PolyMath High across ten languages grouped by resource level.
Group and Overall averages are mean performance within each group and across all ten languages, respectively.
Bold and underlining indicate the best and second-best scores.
}
\label{tab:main_results}
\end{table*}
\begin{table*}[t]
\centering
\small
\setlength{\tabcolsep}{7pt}
\begin{tabular}{llcccc|c}
\toprule
\multirow{2}{*}{Prefix Selection}
& \multirow{2}{*}{Recoverability Filter}
& \multicolumn{4}{c|}{Avg@3 (\%)}
& \multirow{2}{*}{$\Delta_{\mathrm{COPSD}}$} \\
\cmidrule(lr){3-6}
& & Low & Medium & High & Overall & \\
\midrule

Full trajectory (COPSD)
& None
& 1.87 & 8.89 & 16.80 & 9.95 & -- \\

\midrule

\multirow{3}{*}{Teachable prefix}
& None (all)
& 2.40 & 10.53 & 17.20 & 10.76 & +0.81 \\

& Failed only
& 1.96 & 9.87 & 16.67 & 10.21 & +0.26 \\

\rowcolor{oursblue}
\cellcolor{white}
& \textbf{Successful only (Ours)}
& \textbf{2.76} & \textbf{11.02}
& \textbf{18.67} & \textbf{11.60}
& \textbf{+1.65} \\

\bottomrule
\end{tabular}
\caption{
\textbf{Ablation of Teachable Prefix Selection and Teacher Recoverability Verification.}
Avg@3 (\%) on PolyMath High with Qwen3-1.7B.
$\Delta_{\mathrm{COPSD}}$ denotes the overall improvement over COPSD in percentage points.
}
\label{tab:recovery_ablation}
\end{table*}
\begin{figure*}[t]
    \centering
    \includegraphics[width=1.0\textwidth]{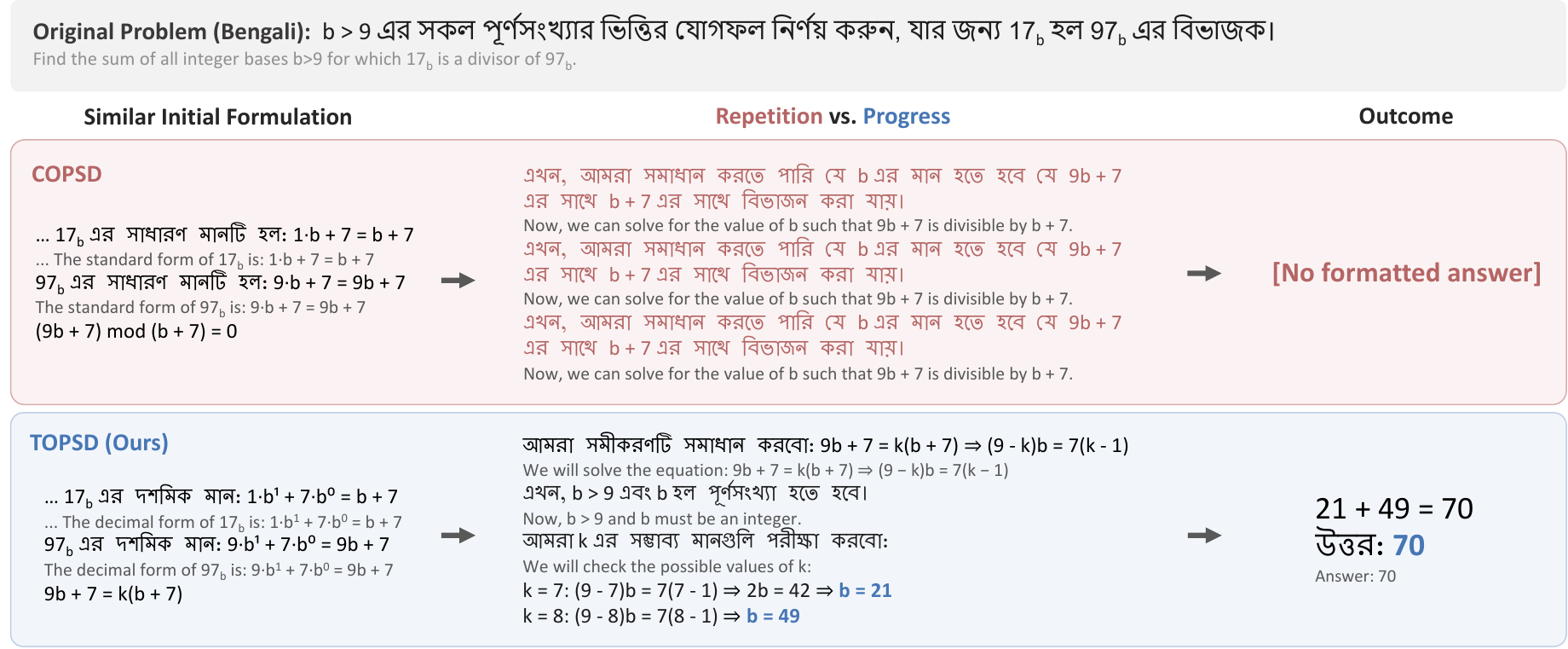}
    \caption{
        \textbf{Qualitative comparison of COPSD and TOPSD on Bengali mathematical reasoning.}
        Starting from similar correct formulations, COPSD becomes stuck in repeated divisibility statements, while TOPSD continues the derivation to $b=21$ and $b=49$ and obtains the correct answer $70$.
        English translations are shown in gray.
    }
    \label{fig:main_qualitative}
\end{figure*}

\paragraph{Dataset and Languages.}
We use the multilingual OpenThoughts dataset released by COPSD~\citep{liu2026crosslingual}, sampling 500 training instances per language (5,000 in total).
We consider ten languages: Bengali (BN), English (EN), Spanish (ES), Japanese (JA), Korean (KO), Russian (RU), Swahili (SW), Telugu (TE), Thai (TH), and Chinese (ZH).

\paragraph{Models and Baselines.}
We conduct experiments using Qwen3-1.7B, Qwen3-4B, and Qwen3-8B~\citep{yang2025qwen3}.
The student is trained with LoRA~\citep{hu2022lora}, while the privileged teacher is the frozen base model conditioned on the English problem and reference solution.
We compare TOPSD against the base model (Vanilla), OPSD~\citep{zhao2026self}, and COPSD~\citep{liu2026crosslingual}.

\paragraph{Training and Evaluation.}
Following OPSD ~\citep{zhao2026self} and COPSD~\citep{liu2026crosslingual}, we use a fixed privileged teacher and full-vocabulary logit distillation.
We evaluate multilingual mathematical reasoning on the PolyMath High split~\citep{wang2026polymath}, reporting Avg@3 and Pass@3 across ten languages.
Detailed training configurations and evaluation protocols are provided in Appendix~\ref{sec:training_details}.

\subsection{Main Results}
\textbf{Overall Performance.}
Table~\ref{tab:main_results} reports Avg@3 and Pass@3 on PolyMath High across three Qwen3 model scales.
TOPSD achieves the best overall performance for both metrics at every scale.
For Avg@3, TOPSD reaches 11.60\%, 17.79\%, and 25.12\% with Qwen3-1.7B, 4B, and 8B, improving over the strongest baseline by 1.65, 1.07, and 2.64 percentage points, respectively.
The corresponding Pass@3 scores are 18.72\%, 26.00\%, and 35.28\%, with gains of 2.08, 0.88, and 3.36 points.
These consistent improvements indicate that selective teacher supervision remains effective as model scale increases.

\paragraph{Performance Across Language Resource Levels.}
The gains are most consistent in low- and medium-resource languages, where TOPSD achieves the best Avg@3 and Pass@3 at all three model scales.
Relative to the strongest baseline, Avg@3 improves by 0.89-3.47 percentage points and Pass@3 by 0.80-4.00 points across these groups.
The gains are largest for Qwen3-8B in the medium-resource group, improving Avg@3 from 24.09\% to 27.56\% and Pass@3 from 35.73\% to 39.73\%.
Overall, TOPSD delivers its strongest and most consistent improvements in low- and medium-resource languages while remaining competitive in high-resource languages.

\subsection{Ablation Studies}
\paragraph{Is Teachable Prefix Selection Alone Sufficient?}
We investigate whether Teachable Prefix Selection alone provides reliable supervision or whether Teacher Recoverability Verification is also necessary.
Table~\ref{tab:recovery_ablation} compares COPSD with three variants that supervise all selected prefixes, only failed-recovery prefixes, or only successful-recovery prefixes (TOPSD).
Prefix selection alone improves Avg@3 from 9.95\% to 10.76\%, while verification further increases it to 11.60\%.
In contrast, supervising failed-recovery prefixes yields lower performance than supervising all selected prefixes or successful-recovery prefixes, highlighting the importance of verifying teacher recoverability.

\subsection{Additional Analysis}
\paragraph{Teachable Prefix Selection Across Model Scales.}
\begin{table}[t]
    \centering
    \small
    \setlength{\tabcolsep}{4pt}
    \begin{tabular}{lccccc}
    \toprule
    \multirow{2}{*}{Model}
    & \multirow{2}{*}{\shortstack{Cutoff \\Detection \\Rate (\%)}}
    & \multicolumn{4}{c}{Relative Cutoff Position (\%)} \\
    \cmidrule(lr){3-6}
    & & Low & Medium & High & Overall \\
    \midrule
    Qwen3-1.7B & 95.94 & 4.09 & 13.65 & 13.02 & 10.47 \\
    Qwen3-4B   & 98.60 & 5.72 & 9.24  & 8.00  & 7.69  \\
    Qwen3-8B   & 99.28 & 6.36 & 10.19 & 5.03  & 6.98  \\
    \bottomrule
    \end{tabular}
    \caption{
    \textbf{Teachable prefix selection statistics.}
    Cutoff Detection Rate is the percentage of trajectories with a detected cutoff.
    Relative Cutoff Position reports the cutoff as a percentage of trajectory length, averaged over trajectories with detected cutoffs.
    Overall is the mean relative cutoff position over all trajectories with detected cutoffs across the ten languages.
    }
    \label{tab:drift_point_stats}
\end{table}
Table~\ref{tab:drift_point_stats} analyzes the selected cutoff positions across model scales and language-resource levels.
Valid cutoffs are detected in over 95\% of trajectories across all model sizes, with average relative positions ranging from 6.98\% to 10.47\%.
The cutoff positions also vary across resource levels, generally occurring earlier in low-resource languages for Qwen3-1.7B and 4B.
These results suggest that teachable prefix lengths depend on both the language and model scale, supporting trajectory-specific prefix selection.

\paragraph{Generalization Beyond Multilingual Reasoning.}
To examine whether TOPSD generalizes beyond multilingual settings, we additionally evaluate it on the English mathematical reasoning benchmarks used in the original OPSD study~\citep{zhao2026self}.
We train Qwen3-4B on 500 English mathematical reasoning problems sampled from OpenThoughts~\cite{guha2025openthoughts} for 160 optimization steps and compare Vanilla, OPSD, and TOPSD. % ($K=4$, $\tau=0.9$).
We evaluate on AIME 2024~\citep{aime24} and AIME 2025~\citep{aime25}, reporting Avg@3 and Pass@3.

As shown in Table~\ref{tab:english_math}, TOPSD achieves the highest Pass@3 on AIME 2024 and improves both Avg@3 and Pass@3 on AIME 2025.
These results suggest that the benefits of TOPSD can extend beyond multilingual reasoning to challenging English mathematical reasoning tasks.

\subsection{Qualitative Results}
To examine reasoning progression beyond final-answer accuracy, Figure~\ref{fig:main_qualitative} compares COPSD and TOPSD on a Bengali mathematical reasoning problem asking for the sum of all integer bases $b>9$ such that $17_b$ divides $97_b$.
Both models begin from a similar correct formulation, expressing the two numbers as $b+7$ and $9b+7$ and establishing the corresponding divisibility condition.
COPSD then repeatedly restates that $9b+7$ must be divisible by $b+7$, without deriving further constraints on $b$, and fails to produce a formatted final answer.
In contrast, TOPSD introduces an integer quotient $k$ through $9b+7=k(b+7)$ and rearranges the equation into $(9-k)b=7(k-1)$.
It then identifies the valid bases $b=21$ and $b=49$, corresponding to $k=7$ and $k=8$, respectively, and obtains their sum, $70$.
This example illustrates how the two models diverge after a correct initial setup: COPSD becomes trapped in repetition, whereas TOPSD develops the divisibility condition into concrete constraints and completes the solution.
\section{Conclusion}
\begin{table}[t]
    \centering
    \small
    \setlength{\tabcolsep}{6pt}
    \begin{tabular}{lcc}
    \toprule
    Method
    & \shortstack{AIME24 \\ \scriptsize Avg@3 / Pass@3}
    & \shortstack{AIME25 \\ \scriptsize Avg@3 / Pass@3} \\
    \midrule

    \textbf{Qwen3-4B} Vanilla
    & \textbf{76.67} / 80.00
    & 64.44 / 76.67 \\

    \quad OPSD
    & \textbf{76.67} / 86.67
    & 67.78 / 83.33 \\

    \rowcolor{oursblue}
    \quad \textbf{TOPSD}
    & \textbf{76.67} / \textbf{90.00}
    & \textbf{71.11} / \textbf{86.67} \\

    \bottomrule
    \end{tabular}
    \caption{
    \textbf{Generalization to English mathematical reasoning} on AIME 2024 and AIME 2025.
    Results are reported in percentages.
    Bold indicates the best result for each metric.
    }
    \label{tab:english_math}
    \vspace{10pt}
\end{table}
We identify \emph{Teacher Drift}, a failure mode in which a privileged teacher increasingly follows erroneous student reasoning as errors accumulate in the prefix. 
To address this, we propose Teachable On-Policy Self-Distillation (TOPSD), which combines Teachable Prefix Selection and Teacher Recoverability Verification to determine when teacher supervision remains appropriate and whether it should be trusted. 
Experiments on PolyMath High across ten languages and three Qwen3 model scales show that TOPSD consistently outperforms OPSD and COPSD, with particularly strong gains in low- and medium-resource languages. 
These results demonstrate that reliable on-policy self-distillation requires not only a strong teacher, but also knowing when the teacher remains able to teach and whether its guidance should be trusted.
Extending this approach beyond mathematical reasoning is a promising direction for future work.
\section*{Limitations}
Our findings are limited to mathematical reasoning, and their generalization to other reasoning domains and model families remains to be explored. 
In multilingual settings, the usefulness of teacher supervision may depend on the model's existing capabilities in each language, so the extent of improvement may vary across languages and models. 
Our current setup also uses English reference solutions as privileged information, leaving other source languages and settings without such resources unexplored. 
Future work could investigate how the language and quality of reference solutions affect supervision and evaluate the approach across a broader range of multilingual reasoning tasks.

\section*{Ethical Considerations}
\paragraph{AI assistance.}
AI tools were used to assist with drafting and revising the manuscript.
The authors retain responsibility for all content, results, and claims.

\paragraph{Potential risks.}
TOPSD improves multilingual mathematical reasoning, while performance differences across languages remain.
Models trained with TOPSD may inherit biases and errors from their underlying models and training data, which can affect the quality of assistance available to speakers of underrepresented languages.
We report results for individual languages and resource groups to make these differences visible.
Applications should be evaluated in their intended languages and contexts, with attention to whether aggregate improvements translate into benefits for the users they serve.
Training and teacher verification also consume computational resources and have associated environmental costs.
We document our training configurations and hardware to support reproducibility and assessment of computational requirements.
These costs should be considered alongside performance gains when scaling the approach to additional models and languages.

\FloatBarrier
\bibliography{main}

\clearpage
\appendix
\section{Additional Experimental Details}
\paragraph{Training Details.}
\label{sec:training_details}

We use Qwen3-1.7B, Qwen3-4B, and Qwen3-8B as the underlying models for both the student and the privileged teacher~\cite{yang2025qwen3}.
The student is trained with LoRA~\citep{hu2022lora}, while the privileged teacher is obtained by disabling the LoRA adapters and freezing the underlying base model. 
Following the COPSD~\cite{liu2026crosslingual} configuration, we use LoRA with a rank of 64 and a scaling factor of 128. The adapters are applied to all attention and MLP projection layers in each Transformer block, including \texttt{q\_proj}, \texttt{k\_proj}, \texttt{v\_proj}, \texttt{o\_proj}, \texttt{gate\_proj}, \texttt{up\_proj}, and \texttt{down\_proj}. 
We do not apply LoRA to the token embeddings, language-model head, or normalization layers.

For each training instance, we sample one on-policy trajectory from the student with a maximum completion length of 2,048 tokens and a sampling temperature of 1.1.
We train only the LoRA parameters for 160 optimization steps using AdamW~\cite{loshchilov2019decoupled} with bfloat16 precision, a learning rate of $1\times10^{-6}$, gradient clipping of $0.1$, and an effective prompt batch size of 32. 
TOPSD performs full-vocabulary logit distillation using the fixed privileged teacher.

Unless otherwise specified, TOPSD uses a likelihood-ratio threshold of $\tau=0.9$ and a persistence length of $K=4$.
For Teacher Recoverability Verification, the privileged teacher greedily generates a continuation from the selected prefix.
Its continuation is evaluated using a rule-based verifier that extracts the final answer from \verb|\boxed{}| and compares it with the ground-truth answer.
Only trajectories that pass verification contribute to distillation, with the loss averaged over supervised tokens per trajectory and balanced equally across active languages.

\paragraph{Implementation Details.}
\begin{table}[t]
\centering
\small
\setlength{\tabcolsep}{4.5pt}
\renewcommand{\arraystretch}{1.08}
\begin{tabular}{@{}>{\raggedright\arraybackslash}p{0.53\linewidth}>{\raggedright\arraybackslash}p{\dimexpr0.47\linewidth-2\tabcolsep\relax}@{}}
\toprule
\textbf{Hyperparameter} & \textbf{Value} \\
\midrule
Base models
& Qwen3-1.7B, 4B, 8B \\

Optimizer
& AdamW \\

Training precision
& bfloat16 \\

Learning rate
& $1\times10^{-6}$ \\

Effective batch size
& 32 \\

Training steps
& 160 \\

Gradient clipping
& 0.1 \\

LoRA rank
& 64 \\

LoRA alpha
& 128 \\

LoRA target modules
& Attention and MLP projections \\

Student rollouts per prompt
& 1 \\

Maximum completion length
& 2,048 \\

Sampling temperature
& 1.1 \\

Distillation vocabulary
& Full vocabulary \\

Teacher policy
& Fixed \\

Teacher verification decoding
& Greedy Decoding\\

Likelihood-ratio threshold $\tau$
& 0.9 \\

Persistence length $K$
& 4 \\

Checkpoint interval
& 20 steps \\

Hardware
& 4$\times$RTX A6000 or 4$\times$H200 \\

Teacher verification max. continuation length
& 8,192 (1.7B, 4B); 12,288 (8B) \\
\bottomrule
\end{tabular}
\caption{
Training hyperparameters used for TOPSD. Unless otherwise
specified, the same configuration is used across all model
scales.
}
\label{tab:training_details}
\end{table}
All experiments are conducted using either four NVIDIA RTX A6000 GPUs or four NVIDIA H200 GPUs. 
For Teacher Recoverability Verification, the privileged teacher greedily generates a continuation from the selected student prefix, with a maximum continuation length of 8,192 tokens for Qwen3-1.7B and Qwen3-4B and 12,288 tokens for Qwen3-8B.
We evaluate the resulting trajectory using a rule-based verifier that extracts the final answer from \texttt{\textbackslash boxed\{\}} and compares it with the ground-truth answer.
Under the default TOPSD configuration with Qwen3-1.7B, $K=4$, and $\tau=0.9$, a full 160-step training run takes approximately 9 hours on four NVIDIA RTX A6000 GPUs.
We save checkpoints every 20 optimization steps.
The main training hyperparameters are summarized in Table~\ref{tab:training_details}.

\paragraph{Dataset and Languages}
We consider ten languages: Bengali (BN), English (EN), Spanish (ES), Japanese (JA), Korean (KO), Russian (RU), Swahili (SW), Telugu (TE), Thai (TH), and Chinese (ZH).
We use the multilingual OpenThoughts training set released with COPSD~\citep{liu2026crosslingual}, which contains OpenThoughts~\citep{guha2025openthoughts} problems translated into the target languages. 
Using a fixed random seed of 42, we sample 500 training instances for each of the ten languages, resulting in 5,000 language-specific training instances in total. 
A single model is jointly trained on the combined multilingual training set.
For resource-level analysis, we group these languages into three tiers based on prior resource-level taxonomies~\citep{zhao2026less} and the multilingual reasoning performance of the base model: High (RU, EN, ZH, ES), Medium (JA, KO, TH), and Low (BN, TE, SW).

\paragraph{Prompt Templates}
The student receives only the problem written in the target language. 
The privileged teacher receives the same target-language problem together with the corresponding English problem and English reference solution.
Both the student and the privileged teacher receive a target-language instruction appended to the user message, asking them to reason and provide the final answer in the target language. 
No separate system message is used.
Generation is initialized with the \texttt{\textless think\textgreater} token followed by a language-specific prefill. 
The exact user messages and language-specific prefills are presented in  Table~\ref{tab:student_prompts}.

\begin{table}[t]
    \centering
    \includegraphics[
        page=1,
        width=\columnwidth
    ]{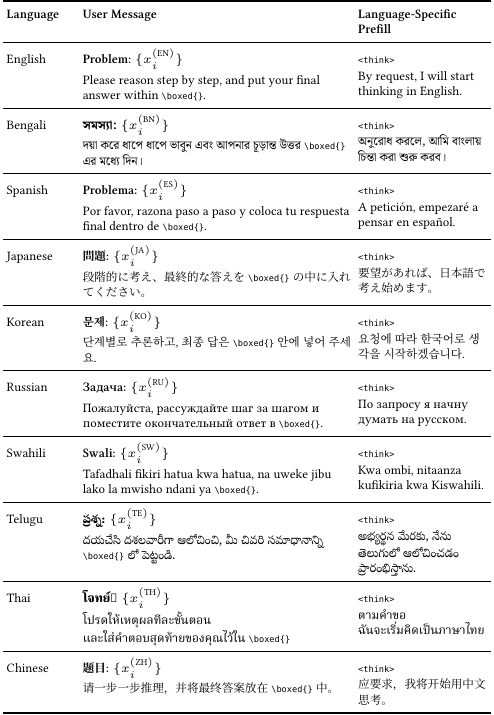}
    \caption{
    Student user messages and language-specific prefills used to initialize multilingual reasoning generation.
    }
    \label{tab:student_prompts}
\end{table}

\begin{table}[t]
    \centering
    \small
    \setlength{\tabcolsep}{3pt}
    \begin{tabular}{@{}lcccc@{}}
    \toprule
    Method & Low & Medium & High & Overall \\
    \midrule

    \multicolumn{5}{l}{\textit{Avg@3 (\%)}} \\
    Fixed Prefix (10\%)
    & 2.22 & 9.42 & 18.00 & 10.69 \\

    \rowcolor{oursblue}
    \textbf{TOPSD}
    & \textbf{2.76} & \textbf{11.02}
    & \textbf{18.67} & \textbf{11.60} \\

    \addlinespace[3pt]
    \multicolumn{5}{l}{\textit{Pass@3 (\%)}} \\
    Fixed Prefix (10\%)
    & 4.80 & 16.27 & 27.80 & 17.44 \\

    \rowcolor{oursblue}
    \textbf{TOPSD}
    & \textbf{5.33} & \textbf{18.67}
    & \textbf{28.80} & \textbf{18.72} \\

    \bottomrule
    \end{tabular}
    \caption{
    \textbf{Comparison with fixed-prefix truncation.}
    Avg@3 and Pass@3 accuracy (\%) on PolyMath High
    using Qwen3-1.7B.
    The fixed-prefix baseline retains the first 10\%
    of each student trajectory, while TOPSD adaptively
    selects teachable prefixes.
    Both methods use the same Teacher Recoverability
    Verification procedure.
    Bold indicates the best result for each metric.
    }
    \label{tab:fixed_prefix_comparison}
\end{table}

\paragraph{Evaluation Details.}
We evaluate all models on the PolyMath High split~\cite{wang2026polymath} in ten languages: BN, EN, ES, JA, KO, RU, SW, TE, TH, and ZH.
For each problem, we sample three completions.
Avg@3 measures the average correctness across these completions, while Pass@3 measures whether at least one completion is correct.
During evaluation, the model receives only the problem in the target language and uses the same language-specific prefill as during training. We use a maximum completion length of 8,192 tokens for Qwen3-1.7B and Qwen3-4B, and 12,288 tokens for Qwen3-8B to accommodate its longer 
\begin{breakablealgorithm}
{\\Teachable On-Policy Self-Distillation (TOPSD)}
\label{alg:topsd}
\scriptsize
\begin{algorithmic}[1]
\Require Training batch
$\mathcal{B}=\{(x_i^{(L)},y_i^\star,a_i,\ell_i)\}_{i=1}^{B}$
\Require Student $p_{\mathrm{S},\theta}$,
frozen teacher $p_{\mathrm{T}}$, verifier $V$
\Require Threshold $\tau$ and persistence length $K$

\For{$i=1$ to $B$}

    \State Sample
    $\hat{y}_i^{(L)}
    \sim p_{\mathrm{S},\theta}(\cdot\mid x_i^{(L)})$
    \State $T_i\gets|\hat{y}_i^{(L)}|$

    \Statex \hspace{\algorithmicindent}
    \textit{Teachable prefix selection}

    \For{$t=1$ to $T_i$}
        \State Evaluate $p_{\mathrm{S},i,t}$ and
        $p_{\mathrm{T},i,t}$ on $\hat{y}_{i,<t}^{(L)}$
        \State
        $N_{i,t}
        \gets
        \log p_{\mathrm{T},i,t}(\hat{y}_{i,t}^{(L)})
        -
        \log p_{\mathrm{S},i,t}(\hat{y}_{i,t}^{(L)})$
    \EndFor

    \State
    $\displaystyle
    C_{i,t}\gets
    \exp\!\left(
        \frac{1}{t-1}\sum_{s=1}^{t-1}N_{i,s}
    \right),
    \quad t=2,\ldots,T_i$

    \State
    $\begin{aligned}[t]
        &\mathcal{D}_i \gets
        \bigl\{\,j\in\mathbb{Z}\mid
        2\le j\le T_i-K+1,\\
        &\qquad C_{i,j+k}>\tau,\;
        \forall k\in\{0,\ldots,K-1\}\,\bigr\}
    \end{aligned}$

    \If{$\mathcal{D}_i=\emptyset$}
        \State $d_i\gets T_i+1$
    \Else
        \State $d_i\gets\min\mathcal{D}_i$
    \EndIf

    \State
    $M_{i,t}^{\mathrm{prefix}}
    \gets\mathbb{I}[t<d_i],
    \quad t=1,\ldots,T_i$

    \Statex \hspace{\algorithmicindent}
    \textit{Teacher recoverability verification}

    \State
    $z_{i,\geq d_i}^{(L)}
    \gets
    \Call{GreedyDecode}{
        p_{\mathrm{T}},
        x_i^{(L)},
        y_i^\star,
        \hat{y}_{i,<d_i}^{(L)}
    }$

    \State
    $G_i\gets
    V\!\left(
        \hat{y}_{i,<d_i}^{(L)}
        \oplus z_{i,\geq d_i}^{(L)},
        a_i
    \right)$

    \State
    $M_{i,t}^{\mathrm{TOPSD}}
    \gets G_i M_{i,t}^{\mathrm{prefix}},
    \quad t=1,\ldots,T_i$

    \State
    $n_i\gets
    \sum_{t=1}^{T_i}M_{i,t}^{\mathrm{TOPSD}}$

    \If{$n_i>0$}
        \Statex \hspace{2\algorithmicindent}
        \textit{Distill on original student states}
        \State
        $\displaystyle
        \mathcal{L}_i
        \gets
        \frac{1}{n_i}
        \sum_{t=1}^{T_i}
        M_{i,t}^{\mathrm{TOPSD}}
        D_{\mathrm{KL}}
        \left(
            p_{\mathrm{S},i,t}
            \middle\|
            \operatorname{sg}[p_{\mathrm{T},i,t}]
        \right)$
    \EndIf

\EndFor

\Statex \textit{Language-balanced aggregation}

\State
$\mathcal{B}_{\ell}^{+}
\gets
\{i\in\{1,\ldots,B\}\mid \ell_i=\ell,\;n_i>0\}$

\State
$\mathcal{A}
\gets
\{\ell\mid |\mathcal{B}_{\ell}^{+}|>0\}$

\If{$\mathcal{A}=\emptyset$}
    \State \Return $0$ \Comment{Skip the student update}
\EndIf

\State
$\displaystyle
\mathcal{L}_{\mathrm{TOPSD}}
\gets
\frac{1}{|\mathcal{A}|}
\sum_{\ell\in\mathcal{A}}
\frac{1}{|\mathcal{B}_{\ell}^{+}|}
\sum_{i\in\mathcal{B}_{\ell}^{+}}
\mathcal{L}_i$

\State \Return $\mathcal{L}_{\mathrm{TOPSD}}$

\end{algorithmic}
\end{breakablealgorithm}
\noindent reasoning trajectories. All models are decoded with a temperature of 1.0 and top-$p$ of 0.95, using a maximum context length of 20,000 tokens.
We use a rule-based verifier that extracts the final answer from \verb|\boxed{}| and compares it with the ground-truth answer. 

\section{Additional Method Details}
\label{sec:detail}

\begin{figure*}[t!]
    \centering
    \includegraphics[
        width=\textwidth
    ]{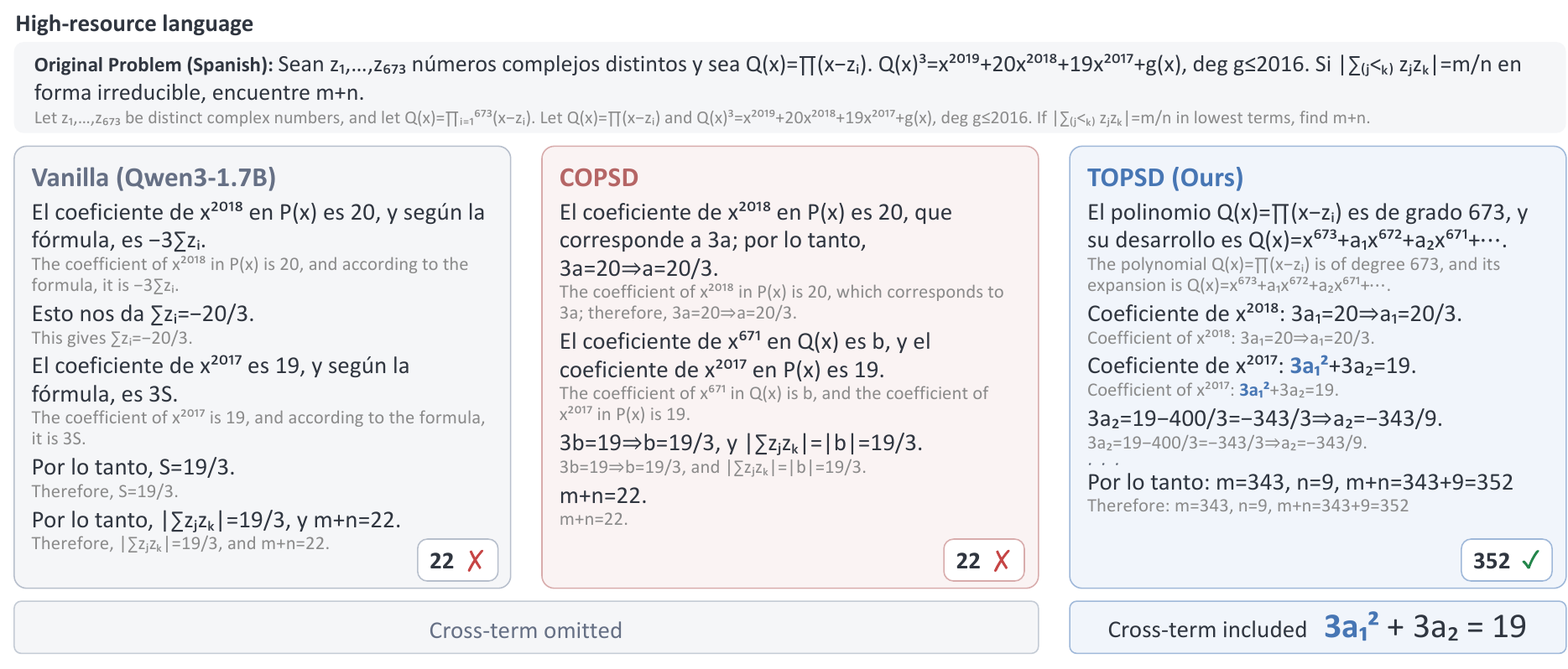}
    \caption{
    \textbf{Qualitative comparison on a high-resource Spanish problem.}
    Both the base model and COPSD omit the cross-term $3a_1^2$ and incorrectly use $3a_2=19$, producing the answer 22. TOPSD correctly matches the coefficient using $3a_1^2+3a_2=19$ and obtains the correct answer 352.
    English translations are shown in gray.
    }
    \label{fig:qualitative_high}
\end{figure*}
\begin{figure}[ht]
    \centering
    \includegraphics[
        width=\columnwidth
    ]{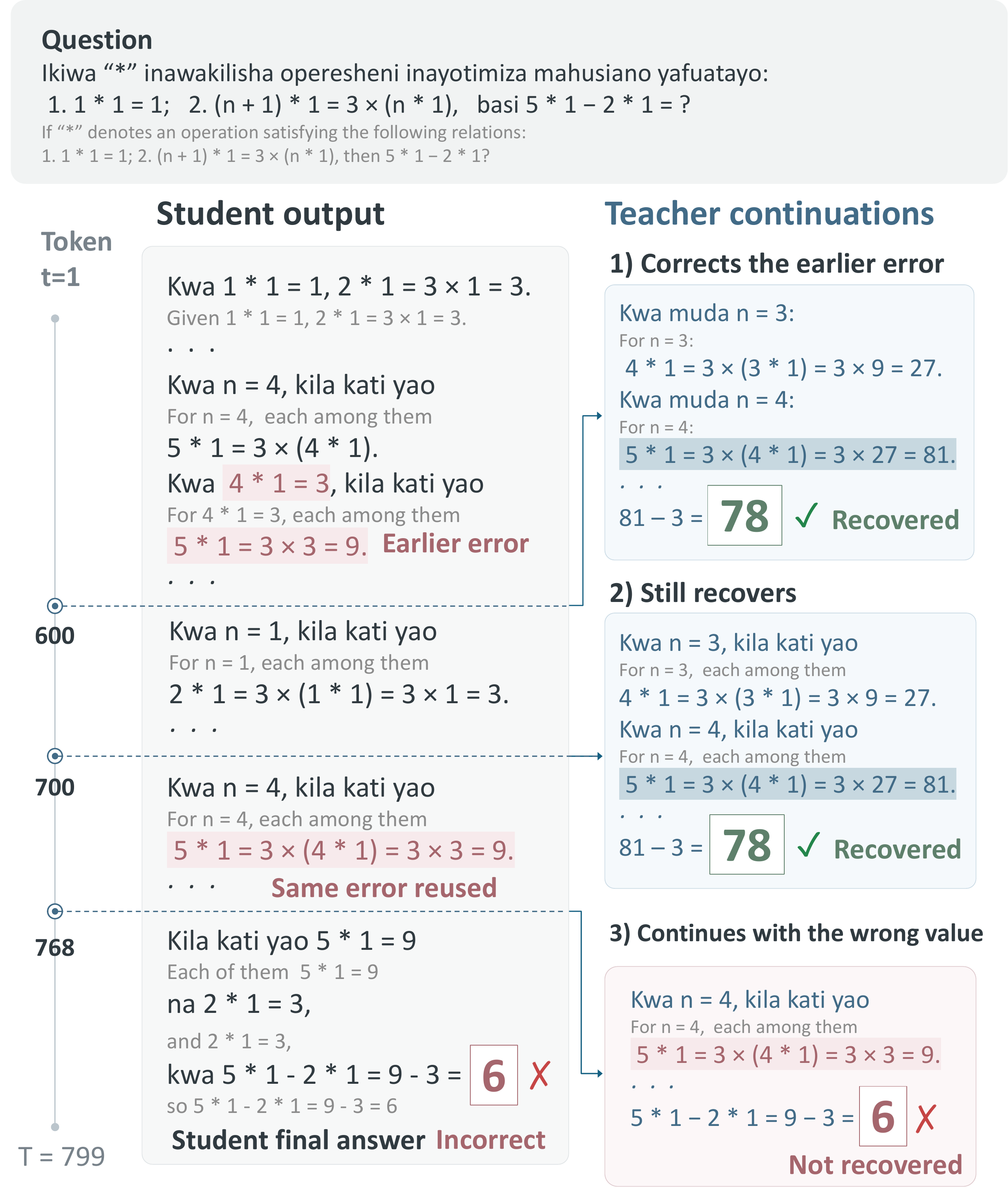}
    \caption{
    \textbf{Completion-based illustration of Teacher Drift.}
    The same privileged teacher is greedily continued from three progressively longer prefixes of a single student completion.
    It recovers the correct answer 78 at $t=600$ and $t=700$, but follows the student's incorrect reasoning and returns 6 at $t=768$.
    This provides an outcome-level view of how prefix accumulation affects teacher recovery.
    English translations are shown in gray.
    }
    \label{fig:additional_teacher_drift}
\end{figure}
\begin{figure*}[ht]
    \centering
    \includegraphics[
        width=\textwidth
    ]{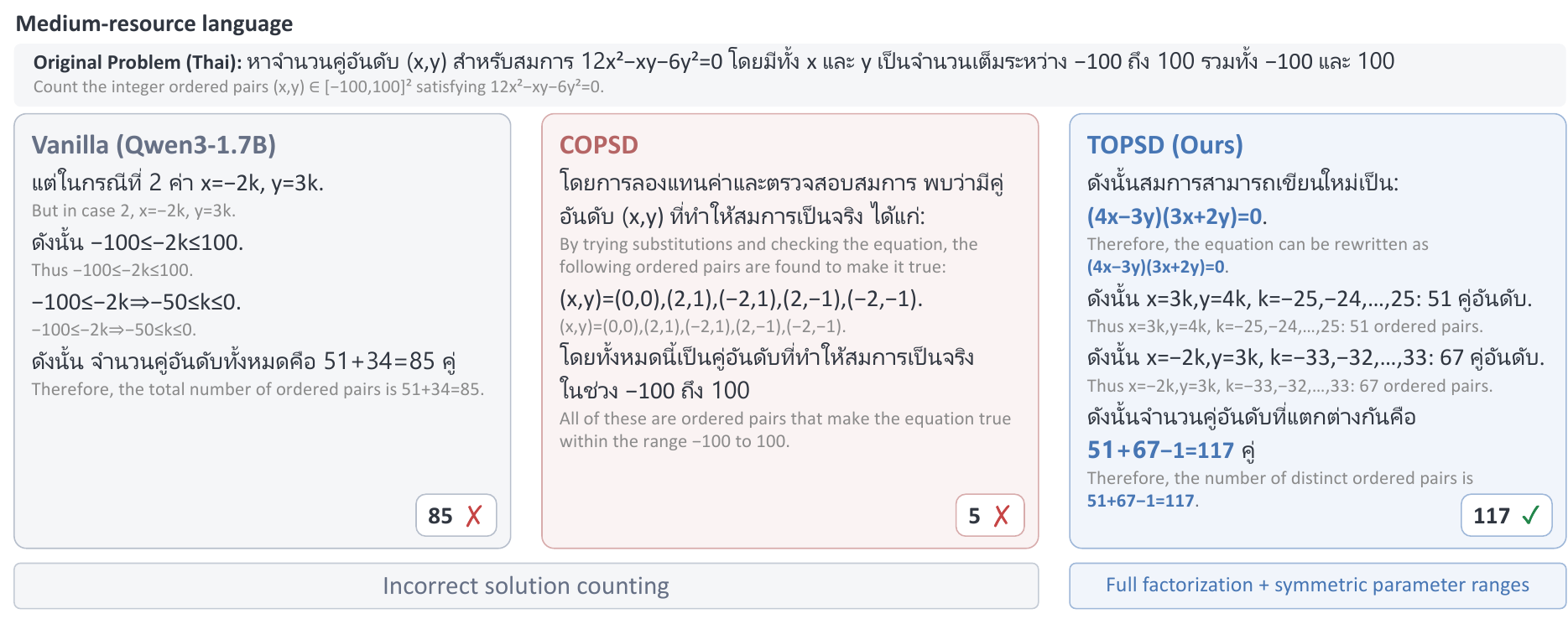}
    \caption{
    \textbf{Qualitative comparison on a medium-resource Thai problem.}
    The base model uses an incorrect asymmetric parameter range, while COPSD reports only an incomplete enumeration of five solutions.
    TOPSD factorizes the equation as $(4x-3y)(3x+2y)=0$, counts both parameterized solution families over their complete ranges, and obtains the correct answer 117. English translations are shown in gray.
    }
    \label{fig:qualitative_medium}
\end{figure*}
\begin{figure*}[ht]
    \centering
    \includegraphics[
        width=\textwidth
    ]{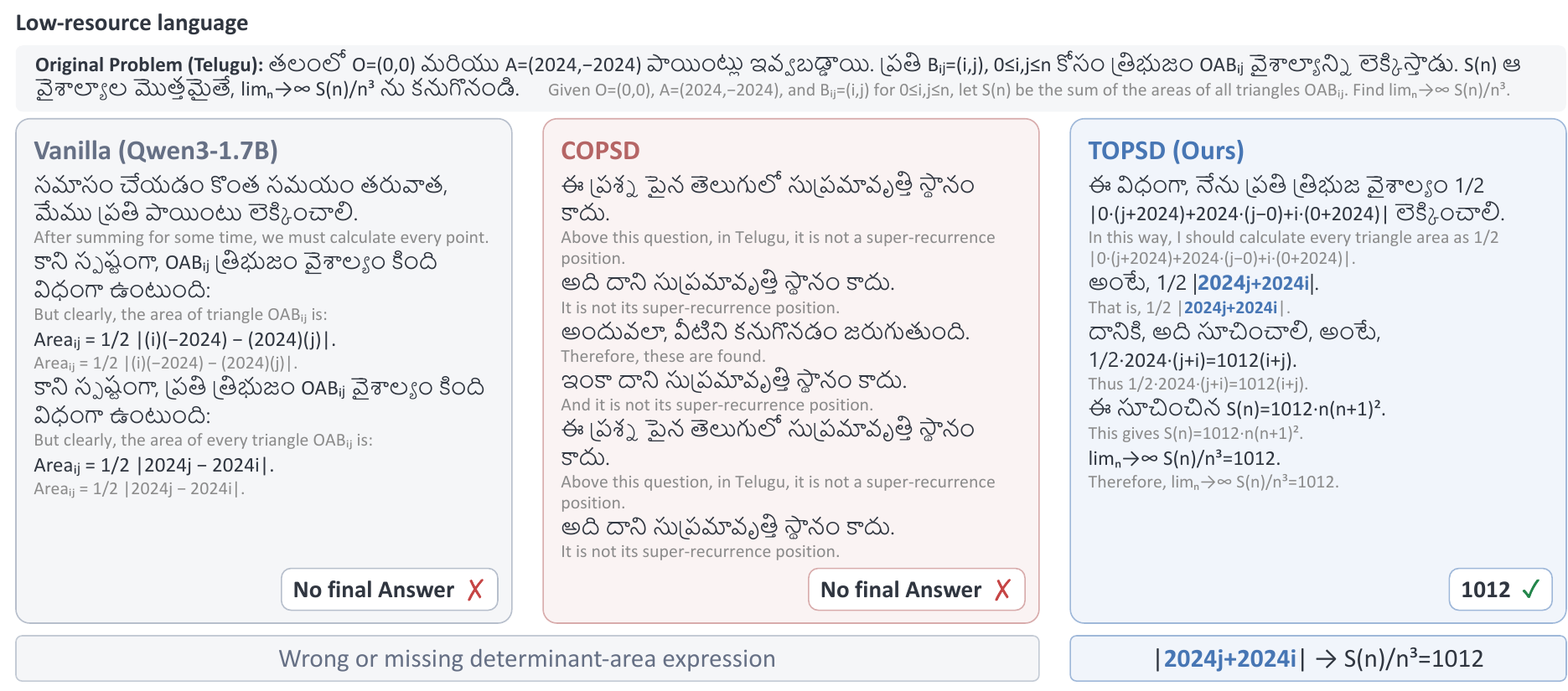}
    \caption{
    \textbf{Qualitative comparison on a low-resource Telugu problem.}
    The base model forms an incorrect determinant-area expression, and COPSD produces repetitive reasoning without a valid area derivation or boxed answer.
    TOPSD correctly derives the triangle area as $1012(i+j)$, computes $S(n)=1012n(n+1)^2$, and obtains the correct limiting value 1012.
    English translations are shown in gray.
    }
    \label{fig:qualitative_low}
\end{figure*}
\begin{table}[t]
\centering
\small
\setlength{\tabcolsep}{4pt}
\begin{tabular}{@{}cccc@{}}
\toprule
$K$
& \shortstack{Cutoff Detection \\ Rate (\%)}
& \shortstack{Relative Cutoff \\ Position (\%)}
& Avg@3 (\%) \\
\midrule
1 & 96.34 & 9.54  & 10.45 \\
2 & 96.22 & 9.87  & 10.94 \\
\rowcolor{oursblue}
\textbf{4} & \textbf{95.94} & \textbf{10.47} & \textbf{11.60} \\
8 & 95.56 & 11.55 & 10.27 \\
\bottomrule
\end{tabular}
\caption{
\textbf{Sensitivity to persistence length $K$ with Qwen3-1.7B.}
Bold marks the default setting ($K=4$).
Cutoff Detection Rate and Relative Cutoff Position characterize the resulting prefix selection, and Avg@3 (\%) is macro-averaged across ten languages on PolyMath High.
}
\label{tab:hyperparameter_sensitivity}
\end{table}
\paragraph{TOPSD Pseudocode}
\label{sec:topsd_pseudocode}
Algorithm~\ref{alg:topsd} presents the complete training
procedure of TOPSD.

\paragraph{Divergence Direction.}
In Eq.~\eqref{eq:opsd}, $D(\cdot,\cdot)$ denotes a general distributional divergence.
Throughout all experiments, we instantiate $D$ as the reverse KL divergence,
\[
D(p_{\mathrm{S}},p_{\mathrm{T}})
=
D_{\mathrm{KL}}
\left(
p_{\mathrm{S}}
\middle\|
\operatorname{sg}[p_{\mathrm{T}}]
\right),
\]
following the COPSD implementation~\cite{liu2026crosslingual}.
Here, $\operatorname{sg}[\cdot]$ denotes the stop-gradient operator.
We use the same divergence direction for all distillation-based methods such as OPSD, COPSD and TOPSD.

\section{Additional Multilingual Results}

\paragraph{Comparison with Fixed-Prefix Truncation.}
To investigate whether the improvements of TOPSD arise simply from restricting distillation to shorter prefixes, we compare our adaptive prefix selection with a fixed-prefix baseline.
Specifically, the fixed-prefix baseline retains the first 10\% of each student-generated trajectory for distillation, regardless of the estimated drift point.
We choose 10\% based on the average relative cutoff position of 10.47\% observed with Qwen3-1.7B under our default configuration ($K=4$, $\tau=0.9$), as reported in Tables~\ref{tab:drift_point_stats} and~\ref{tab:hyperparameter_sensitivity}.

Both methods use the same Teacher Recoverability Verification procedure and training configuration, differing only in their prefix selection strategies.
As shown in Table~\ref{tab:fixed_prefix_comparison}, TOPSD outperforms fixed-prefix truncation, suggesting that its improvements cannot be explained solely by restricting supervision to early reasoning states.
Instead, adaptively selecting teachable prefixes across languages and individual student trajectories provides additional benefits over a fixed cutoff.

\paragraph{Sensitivity to $K$.}
Table~\ref{tab:hyperparameter_sensitivity} examines how the persistence length $K$ affects cutoff selection and downstream performance, with other hyperparameters fixed.
The cutoff detection rate measures the proportion of trajectories with a detected cutoff, while the relative cutoff position denotes the cutoff location normalized by trajectory length.
As $K$ increases, the detection rate decreases slightly, while cutoffs tend to occur later.
This reflects the stricter persistence requirement, which retains longer prefixes for potential supervision.
Avg@3 increases from $K=1$ to $K=4$ but declines at $K=8$, with the default setting ($K=4$) achieving the highest score of 11.60\% among the evaluated values.

\section{Additional Qualitative Analysis}
\label{sec:additional_qualitative_analysis}

\paragraph{Qualitative Comparison of Reasoning Trajectories}
\label{sec:qualitative_comparison}

Figures~\ref{fig:qualitative_high}, \ref{fig:qualitative_medium}, and \ref{fig:qualitative_low} compare the reasoning trajectories produced by the Qwen3-1.7B vanilla model, COPSD, and TOPSD on three multilingual mathematical reasoning problems spanning high-, medium-, and low-resource languages. 
Rather than considering only final-answer correctness, we highlight the local reasoning error that determines each solution.

In the high-resource Spanish example, both the base model and COPSD omit the cross-term $3a_1^2$ when matching the coefficient of $x^{2017}$, whereas TOPSD correctly uses
$3a_1^2+3a_2=19$. 
In the medium-resource Thai example, the base model uses an incorrect parameter range and COPSD reports only an incomplete set of solutions, while TOPSD factorizes the equation and counts both solution families over their full symmetric ranges.
In the low-resource Telugu example, the base model forms an incorrect triangle-area expression and COPSD produces repetitive reasoning without a valid derivation, whereas TOPSD derives the correct determinant-based area and obtains the limiting value.

Across all three examples, the improvement from TOPSD is not limited to changing the final answer. 
TOPSD corrects the key intermediate derivation on which the remainder of the solution depends, resulting in more coherent and complete reasoning trajectories across language-resource levels.

\paragraph{Additional Examples of Teacher Drift.}
While Figure~\ref{fig:teacher_assimilation} of the main paper directly examines the teacher's next-token predictions at distillation states, Figure~\ref{fig:additional_teacher_drift} provides a complementary completion-based view.
We greedily continue the same privileged teacher independently from progressively longer prefixes of a single Swahili student trajectory.

The recurrence implies $2*1=3$, $3*1=9$, $4*1=27$, and $5*1=81$, giving the correct answer $81-3=78$.
However, the student incorrectly substitutes $4*1=3$ into $5*1=3\times(4*1)$, obtaining $5*1=9$ and the wrong answer $9-3=6$.
The red highlights mark this incorrect substitution and its propagation, which reappears later in the student's reasoning.

At $t=600$ and $t=700$, the teacher recovers despite the earlier error already being present in the prefix.
It correctly computes $4*1=27$ and uses this value to obtain $5*1=81$, as highlighted in blue, ultimately returning $78$.
At $t=768$, the teacher instead continues the incorrect derivation and returns $6$, matching the student's wrong answer.
Thus, in this trajectory, the teacher can initially recover from an erroneous prefix but fails to do so after the prefix is extended with further incorrect reasoning.

\end{document}